\documentclass[a4paper,twoside]{article}

\usepackage{booktabs}
\usepackage{epsfig}
\usepackage{subcaption}
\usepackage{calc}
\usepackage{amssymb}
\usepackage{amstext}
\usepackage{amsmath}
\usepackage{amsthm}
\usepackage{multicol}
\usepackage{pslatex}
\usepackage{apalike}
\usepackage{algorithm2e}
\usepackage[bottom]{footmisc}
\usepackage{SCITEPRESS}     

\begin{document}

\title{High Semantic Features for the Continual Learning of Complex Emotions: a Lightweight Solution}

\author{\authorname{Thibault Geoffroy\sup{1}, Gauthier Gerspacher\sup{1} and Lionel Prevost\sup{1}}
\affiliation{\sup{1} EsieaLab - Learning, Data, Robotics, 9 rue Vésale, 75005 Paris, France}
\email{thibault.geoffroy@esiea.fr, gerspacher@et.esiea.fr, lionel.prevost@esiea.fr}
}


\abstract{  
Incremental learning is a complex process due to potential catastrophic forgetting of old tasks when learning new ones. This is mainly due to transient features that do not fit from task to task. In this paper, we focus on complex emotion recognition. First, we learn basic emotions and then, incrementally, like humans, complex emotions. We show that Action Units, describing facial muscle movements, are non-transient, highly semantical features that outperform those extracted by both shallow and deep convolutional neural networks. Thanks to this ability, our approach achieves interesting results when learning incrementally complex, compound emotions with an accuracy of 0.75 on the CFEE dataset and can be favorably compared to state-of-the-art results. Moreover, it results in a lightweight model with a small memory footprint.}

\onecolumn \maketitle \normalsize \setcounter{footnote}{0} \vfill

\section{\uppercase{Introduction}}
\label{sec:intro}

In the fourth industrial age, artificial intelligence (AI) plays a crucial role in many of our activities and endeavors. AI is able to match or even outperform humans in cognitive tasks such as image recognition and natural language processing. But in the uniquely human domains where communication, empathy, and compassion are needed, AI is still limited in handling those more subtle tasks.

Communication between humans is fundamental to our ability to learn, work, and interact. Among communication channels, facial expressions are one of the most efficient vectors to express emotional states and intentions. In fact, more than 55\% of emotional communication (also called non-verbal communication) is conveyed through facial expressions \cite{noword}. As a result, the automatic recognition of facial emotions is becoming a crucial component, as recognizing emotional states forms the foundation for a computer's understanding of emotions and its subsequent responses.

In recent decades, numerous Facial Expression Recognition (FER) approaches have been investigated in the fields of computer vision and machine learning to encode expression information from face representations. Unfortunately, most of these studies focus on basic emotion recognition such as joy, fear, and anger \cite{ekman_basic_1992}. But in real day-to-day situations, recognizing only these basic universal emotions is no longer enough, and recognizing complex emotions becomes a necessity.  For example, in the area of education, when students face problems during pedagogical activities and exhibit complex emotional states like misunderstanding, stress or pride.

Current state of the art in basic emotion recognition primarily relies on deep learning techniques (e.g., Deep Convolutional Neural Networks and, recently, Transformers). In recent years, these methods have enabled computer vision systems to achieve highly efficient results, outperforming classical methods that rely on handcrafted static features. Surprisingly, these methods exhibit poor performance on complex emotion recognition. On the contrary, continual learning methods perform better. 

In this study, we decided to address the problem of complex emotion recognition by incrementally learning these latter. In fact, our aim was to mimic humans. As humans, studies show that we learn to express and recognize basic emotions during our early childhood \cite{denham98} \cite{pons2004} \cite{widen2008}, while complex and subtle emotions are learned continuously throughout our life. The other aim of this study is to promote a lightweight sustainable model, with a low memory and carbon footprint.

The study is divided into two parts. First, we compare the behavior of two network architectures using the same incremental learning strategy found in the literature. The first one is shallow and hand-crafted, while the second is deeper and already pre-trained. We use both as feature extractors from images, make copies of them and train these copies independently, task by task, to build an incremental, multiple expert recognizer (the process is detailed in section \ref{sec:CILNN}). That way, we extract low or mid-level features that are transient since they can change from one task to another. In the second part, we extract highly semantic features (facial muscle activation also known as Facial Action Units) directly from images, using an open-source research tool. Contrary to the previous features, these Action Units are non-transient and stay the same during all the learning process, whatever the emotion and its complexity. 
All these experts also use class-conditional Gaussian models to model class distributions and perform classification.
so, our main contributions in the FER domain are the following:

\begin{itemize}
    \item Features extracted by CNNs, whatever the complexity of the latter and even if it is trained continuously, have a low to mid semantic level. Thus, they cannot discriminate the subtle differences that may appear between complex emotions.
    
    \item Action Units, on the other hand, have a higher semantic level and offer strong discrimination capabilities for complex emotions. Moreover, since they are non-transient features, they do not need continual learning. 
    
    \item Gaussian models constitute a lightweight solution as density estimators and powerful emotion predictors when used on a smart feature space. Moreover, the memory footprint of an ensemble of Gaussian models is really low, compared to neural networks.
\end{itemize}

The paper is organized as follows. Section \ref{sec:SOA}
 presents in detail the different ways of describing a facial emotion and several solutions to train a classifier incrementally. In section \ref{sec:methodo}, we describe the methodology, including pre-processing and emotion recognition using different feature extractors: low-level CNN features and high-level Action Units. Section \ref{sec:expe} is devoted to a thorough analysis of the experimental results. Finally, the last section analyzes the findings and concludes this work. 

\section{\uppercase{State of the art}}
\label{sec:SOA}

\subsection{Basic emotions, complex emotions and action units}
Basic emotions are generally recognized throughout cultures and are commonly linked to specific physical reactions and
facial expressions. Fear, anger, disgust, sadness, joy, and surprise are among the most frequently acknowledged
basic emotions \cite{ekman_basic_1992}.

\begin{figure}[htbp]
    \centering
    \includegraphics[width=0.5\textwidth]{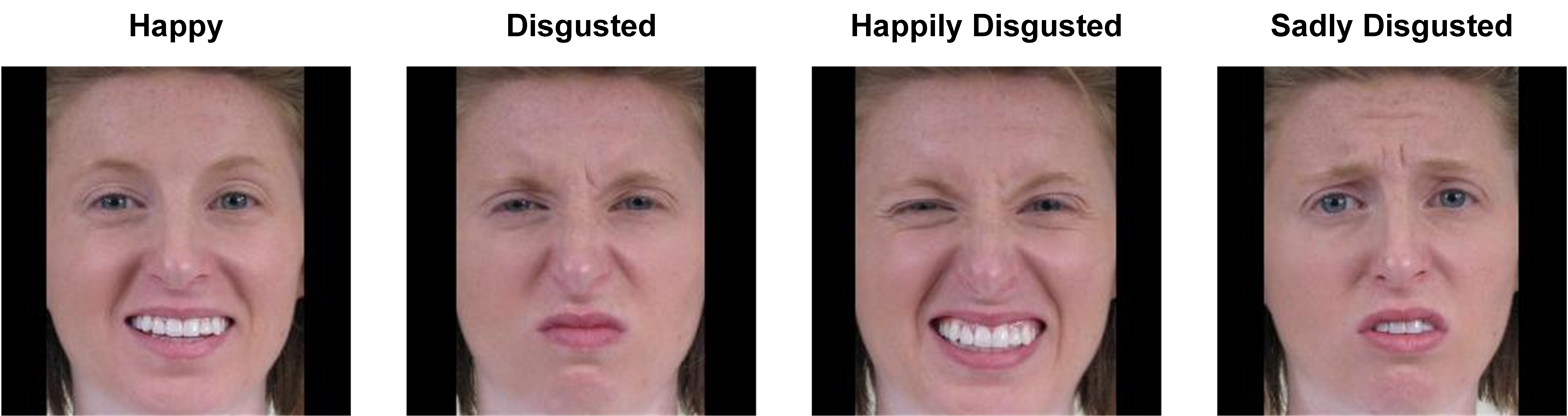}
    \caption{Basic and compound emotions  \cite{maiden2023complexfacialexpressionrecognition}.}
    \label{fig:compound_emo}
\end{figure}

Complex emotions are intricate and multifaceted emotional experiences that encompass a blend of basic emotions,
cognitive processes, and social influences. Contrary to basic emotions that are short-lived, complex emotions tend to persist for extended periods.  Furthermore, they frequently result from combinations or interactions of basic emotions \cite{maiden2023complexfacialexpressionrecognition} and are more likely to involve mixed sentiments or uncertainty. That is why they are usually called compound emotions or expressions. In \cite{cfee}, authors define 21 categories of compound emotions that humans express in-the-wild. For example, jealousy could blend fear (of losing a relationship) with anger (toward a perceived rival). Thus, the corresponding compound emotion would be "fearfully angry". Most of the time, compound expressions are more than the sum of their basic parts, as shown in figure \ref{fig:compound_emo}.

\begin{figure}[htbp]
    \centering
    \includegraphics[width=0.4\textwidth]{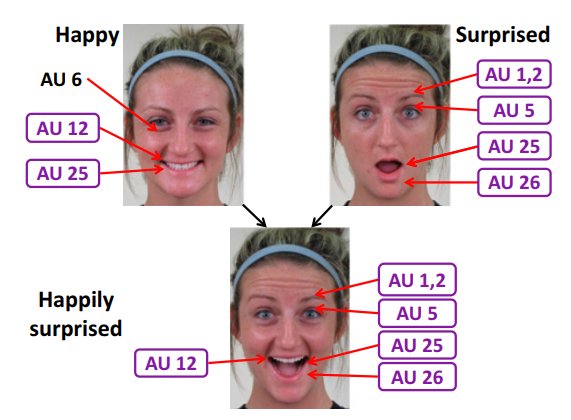}
    \caption{Facial Action Coding \cite{cfee}.}
    \label{fig:FACS}
\end{figure}

The previous methods of expression categorization are highly subjective. In contrast, the Facial Action Coding System (FACS) \cite{facs_1983} offers an objective way to describe expressions. FACS encodes the activation of facial muscles as known as Action Units (AUs). For example, AU1 encodes the raising of the inner brow and AU20, the lip stretcher. Expression of emotion involves several AUs and their intensity indicates the degree of expression of this emotion (see figure \ref{fig:FACS}. It is important to note that a single expression can be produced by various sets of AUs, depending on the subject. Nevertheless, AUs are highly semantic features that could be used to predict emotions with a high accuracy, as demonstrated in \cite{senechal_2014}.

Action Units detection has been extensively studied by the affective computing community over two decades. In summary, methods have evolved (as in many domains in pattern recognition) from handcrafted features (such as Haar or Gabor filters) used to feed a SVM \cite{senechal2011}, to deep CNNs \cite{akay2018}. \cite{zhi2020} presents a detailed study of these methods. Recently, vision transformers have been proposed \cite{yuan2024}.
But most of these methods cannot easily handle  incremental learning and suffer from catastrophic forgetting.

\subsection{Incremental Learning}
Also called Continual Learning or Lifelong Learning, Incremental Learning is a domain of  machine learning  where the data does not arrive in one go but in several phases (also called tasks) and where access to previous data is not allowed. Throughout the Incremental Learning process, the distribution of data points within a class or the number of labels can change. This creates a scenario where machine learning models need to be able to acquire knowledge from new data (i.e. plasticity) while preserving previously gained knowledge (i.e. stability). One of the main pitfalls of Incremental Learning is the tendency of machine learning models to overwrite old knowledge with new, more recent information (lack of stability). This tendency can greatly reduce the performance of the model on previous tasks and is referred to in the literature as catastrophic forgetting \cite{zhou24}. 

Class-Incremental Learning (CIL) represents one of the most challenging scenarios in the field of Incremental Learning. In this scenario, data arrive in batches that contain a number of data points and labels. The model has access to the data of the most recent task for the training.  During inference, Task-Incremental Learning can be done using two different scenarios. The task-agnostic scenario supposes that the task-id is unknown and the classifier has to predict the right label through all the already seen labels. The task-aware scenario supposes that the task id is known during inference.

The easiest solution for CIL is to store exemplars along tasks \cite{wang2022}. But this replay method is not always an option due to memory space and other constraints. Therefore, exemplar-free CIL has been thoroughly studied \cite{Smith21}, \cite{zhu2022}. Several methods exist to avoid replay, such as pseudo-replay of synthetic samples generated by a generative model \cite{shin2017}. Knowledge distillation of older models during the training of new ones to retain knowledge on old tasks (\cite{maiden2023complexfacialexpressionrecognition}, \cite{incactlearn2020}) is also usual. 
Finally, architecture-based methods for CIL can dynamically adjust the network's parameters while learning new tasks. In the extreme case, each task can have a dedicated network as an expert \cite{rypeść2024}. That greatly improves plasticity but also requires more resources as the number of parameters increases. To limit the latter, it is possible to freeze a part of the feature extractors and share it between all experts.

A few years ago, some exemplar-free methods employed nearest mean classifiers with stored class centroids \cite{rebuffi2017}. Recently, in \cite{yang2023}, a fixed, pre-trained feature extractor and class-conditional Gaussian models with diagonal covariance matrices were used to solve CIL problems. In \cite{rypeść2024}, this solution has been extended by building an "expert" which combines a deep network and an ensemble of class-conditional Gaussian models to deal with each task. During CIL, when a new task appears, a new expert inherits the network parameters learned on the previous task. New Gaussian models, conditional to new task classes, are trained. During inference, the log-likelihoods of the sample are computed by all the Gaussian models, and the Bayes rule is used to predict the class. More information on this method will be given in section \ref{sec:CILNN}.

\section{\uppercase{Proposed methodology}}
\label{sec:methodo}

We first describe the image pre-processing (face alignment and AUs extraction). Then, we detail the class-incremental learning algorithm that adds one dedicated expert (shallow or deep CNN) per task before combining them. Finally, we use the same process using AUs and conditional Gaussian Mixture Models (GMM). 

\medskip
\subsection{Pre-processing}
This work mainly uses OpenFace, a tool designed for the affective computing community and researchers interested in building interactive applications based on facial behavior analysis. It is the first non-profit toolkit capable of facial landmark detection, head pose estimation, facial action unit recognition, and eye gaze estimation with available source code \footnote{https://github.com/TadasBaltrusaitis/OpenFace}. The computer vision algorithms that represent the core of OpenFace 2.0 demonstrate state-of-the-art results in all of the above-mentioned tasks \cite{openface}. In detail, landmark detection uses a Constrained Local Model (CLM). Then, face is aligned and encoded by using PCA-reduced Histogram of Oriented Gradient. This vector is concatenated with shape features extracted by the CLM. Finally, a Support Vector Regressor uses this mixed information to predict the intensity of AUs activated in the face.

For training and inference, each image is processed by OpenFace to get a cropped face image, facial landmarks, and a vector containing 17 AUs intensities (on a 0 to 5 scale). Then, the image is aligned using the position of the center of the pupils.

\medskip
\subsection{Incremental learning with Neural Networks}
\label{sec:CILNN}

We propose here an architecture-based incremental learning method based on \cite{rypeść2024} that gets state-of-the-art results on the benchmark dataset CIFAR-100. The core idea of this approach is to directly diversify experts by training them sequentially on different tasks $t$ and then combining their knowledge during the inference. Each expert contains two components: 
\begin{itemize}
    \item A convolutional network that serves as feature extractor and generates a unique feature space.
     \item An ensemble of class-conditional Gaussian models (one per class). 
\end{itemize}

\begin{figure}[h]
    \centering
    \includegraphics[width=0.4\textwidth]{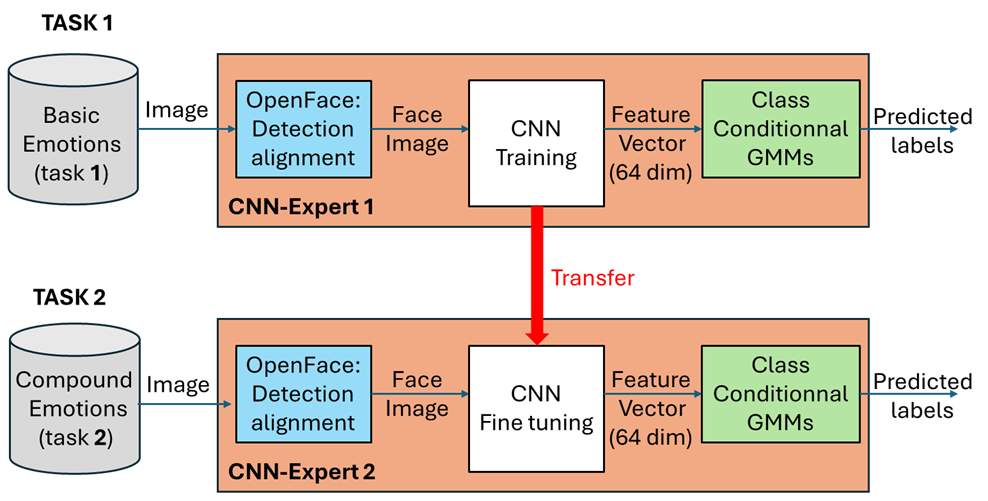}
    \caption{Incremental training of an ensemble of experts (here, number of tasks is T=2).}
    \label{fig:SEED}
\end{figure}

\medskip
\textbf{Algorithm} The problem is divided into two tasks, each corresponding to a set of classes to be learned. As shown in figure \ref{fig:SEED}, an expert is trained on task 1 data. A CNN is learned to discriminate task classes. Then, it is used as an extractor to project images in its own feature space, by deleting dense classification layers. Finally, class-conditional Gaussian models $(\mu_c , \Sigma_c)$ are trained on each class $c$ of task 1, to model class-conditional distributions in the feature space. 
Then, since task 1 has been learned, the corresponding data are erased. 
When task 2 begins, a second CNN is instantiated and inherits the parameters of the first CNN. Transfer learning and fine-tuning are performed, using task 2 data only. Then, Gaussian models are trained on task 2 data, in the new feature space, one for each class. The process is repeated as long as there are tasks left. Considering $T$ tasks, we will finally have  $T$ different experts, each specialized on "their" set of classes, composed of a feature extractor and several Gaussian models. 
This incremental learning algorithm is clearly based on architecture growing and not on replays or regularization methods.
Some tricks are proposed in \cite{rypeść2024} for a large number of classes to constrain growing that are not reported here since the number of classes/tasks stays low.

\begin{figure}[h]
    \centering
    \includegraphics[width=0.4\textwidth]{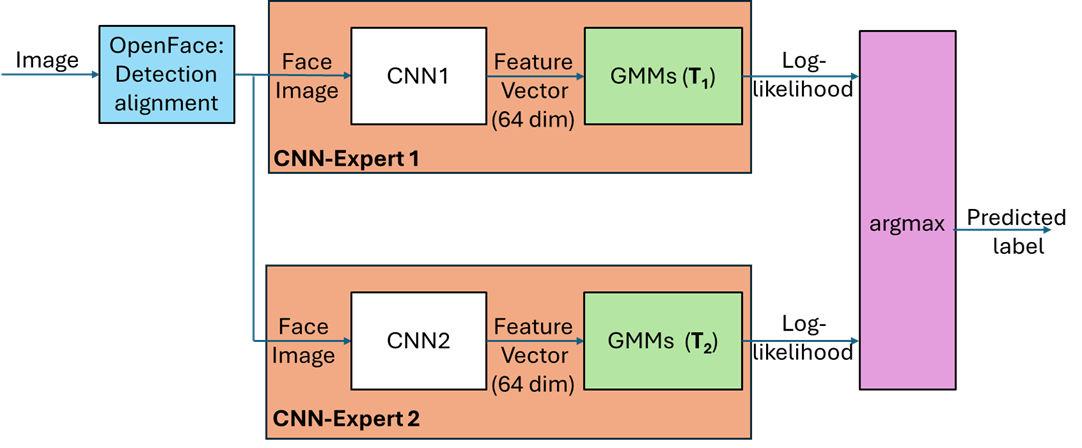}
    \caption{Inference of the ensemble of experts (for T=2).}
    \label{fig:SEED_inference}
\end{figure}

\medskip
\textbf{Inference} To infer, all the experts are activated simultaneously. Each CNN generates a feature representation $r_i$ (the feature vector for task $i$) of the input image $x$ and each Gaussian model calculates its log-likelihoods $L_c$, using the equation below, given the distribution $(\mu_c , \Sigma_c)$ of each class (for each expert separately).

\begin{equation}
L_c(x)=\frac{1}{2}[log\Sigma_c+Slog(2\pi+(r_i-\mu_c)^T\Sigma_c^{-1}(r_i-\mu_c)]
\end{equation}

where $S$ is the size of the feature space. Then, they softmax those log-likelihoods and compute their average over all experts. Finally, Bayes posterior decision rule selects the class with the highest average softmax as the prediction, as shown in figure \ref{fig:SEED_inference}. For task agnostic, all the classes are considered. For task aware, this procedure is limited to classes from the considered task.

\medskip
\textbf{CNN} During incremental learning, a new network $N_t$ is created for each new task $t$. If the task objective is to discriminate $k$ classes, $N_t$ it will have $k$ outputs. For task $t>1$, $N_t$ it inherits the parameters of $N_{t-1}$ and is fine-tuned on task $t$ data.

For feature extraction, we used the last dense layer, which size is 64. In other words, for both training and inference, the input image is processed by the network that outputs a 64-dimensional feature vector.

\medskip
\textbf{GMM} They are used to estimate the underlying distribution of each class given the feature vectors. We use Gaussian Mixture Models instead of simple Gaussian models. We employ diagonal covariance matrices since we do not have enough samples per class to estimate full matrices. Models are trained using the Expectation-Maximization algorithm. The optimal number of components of the mixture is found by using the Akaike Information Criterion. It may vary depending on the heterogeneity of the class.

Each class distribution is modeled by one GMM, trained only on data of this class. Therefore, for $k$ classes and whatever the number of tasks, we will train $k$ GMMs that compute, during inference, log-likelihoods for the input image $x$. We then use the equation below (a variant of Bayes rule) to predict the class $k^*$ of the image $x$.

\begin{equation}
    k^*(x)=argmax_c(L_c(x | \mu_c, \Sigma_c))
\end{equation}

This method is really interesting, as it transforms a set of $k$ GMMs, used as generative models into a discriminative model that separates $k$ classes.

\subsection{Incremental learning with Action Units}

In this section, we use OpenFace to extract 17 AU activations (from a 0 to 5 scale). That way, we no longer need a CNN feature extractor. The new feature vector has a dimension of 17 that is much smaller than before. That is why we can use full covariance matrices in GMMs, hoping for better estimation and higher performance. 

\begin{figure}[htbp]
    \centering
    \includegraphics[width=0.4\textwidth]{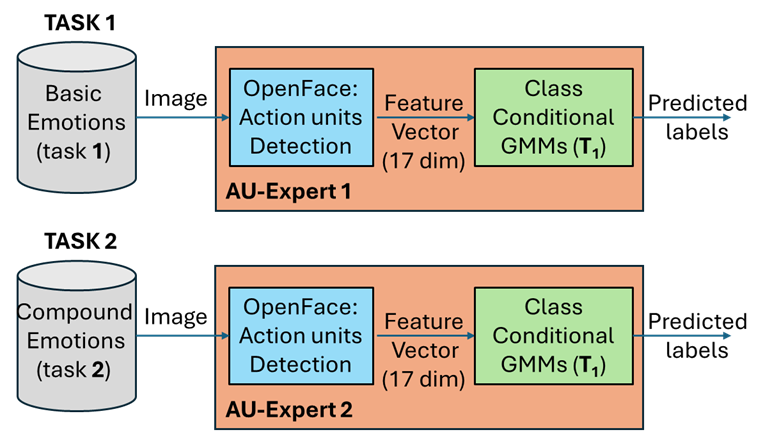}
    \caption{Training an ensemble of AUs based classifiers (for T=2).}
    \label{fig:AU_expert}
\end{figure}

GMMs are learned as previously, from task to task (see figure \ref{fig:AU_expert}). Now, contrary to incremental learning with CNN, there is no longer a need to transfer knowledge from task to task, since the AU detector is already trained efficiently. 
During inference, the GMM that outputs the highest log-likelihood gives the predicted class.

We also evaluate here Bayesian Gaussian Mixture Models (BGMMs) based on variational inference \cite{BGMM} mainly for two reasons. Firstly, they have the ability to avoid singular data point solutions, and, secondly, they can directly determine the optimal number of components without resorting to methods such as cross-validation.

\section{Experimental results}
\label{sec:expe}

\subsection{Dataset and experimental process}
Most of FER databases only include basic emotion images. The Compound Facial Expression of Emotions (CFEE) dataset \cite{cfee} is one of the few public datasets whose purpose is complex emotion recognition. Unfortunately, as far as we know, CASME database \cite{micro-exp} is no longer available. The CMED dataset \cite{CMED} focuses on child micro-expressions, which are not the aim of our study. CFEE contains 5,060 images of 230 subjects, divided into 22 classes: 7 basic emotion classes (including neutral) and 15 compound emotion classes. This distinction between basic and compound emotions leads to structuring the learning process: like humans, machine begins to learn basic emotions and then compound emotions are learned incrementally. Thus, a possible learning scenario is described below:

\begin{itemize}
    \item Task 1: 6 basic emotions and neutral    
    \item Task 2: compound joy (happily surprised, happily disgusted, awed)
    \item Task 3: compound sadness (sadly fearful, sadly angry, sadly surprised, sadly disgusted)
    \item Task 4 compound fear (fearfully angry, fearfully surprised, fearfully disgusted)
    \item Task 5: compound anger (angrily surprised, angrily disgusted, hatred)
    \item Task 6: compound disgust (disgustedly surprised, appalled)
\end{itemize}

Therefore, we have a total of $T=6$ tasks to learn the whole set of compound emotions. Task $t_1$ is devoted to learning basic emotions, while the following tasks $t_2$ to $t_6$ aim to learn a set of compound emotions linked by their primary emotion, like "angrily surprised" and "angrily disgusted". It is important to notice that, due to the incremental learning process, tasks $t_2$ to $t_6$ can be interchanged randomly, \textbf{without impacting} the general performance.

As a performance metric, we use accuracy (ACC), which is the ratio of the correctly classified samples on $t$ tasks (from task $1$ to task $t$) to the total effective of these tasks $n_t$. Given $y_i$ the label and $\hat{y_i}$ the prediction, accuracy is given by:

\begin{equation}
  ACC(t) = \frac{1}{n_t} \sum_{i=1}^{n_t} I(y_i \neq \hat{y_i})
\end{equation}

Regarding state-of-the-art results on the CFEE dataset, we will first detail recent experiments in offline (batch) learning. These are gathered in table \ref{tab:SOA_batch}. 
As far as we know, only three methods have been proposed to learn compound emotions incrementally. Their protocol is not far from ours: they initially learn basic emotions and then, sequentially, compound emotions. Table \ref{tab:SOA_inc} presents their core methods and performances (KD stands for Knowledge Distillation).
When comparing both tables, it is clear that incremental learning performs better than offline methods. This is proof that our proposal (learning incrementally complex emotions) is meaningful.

\begin{table}[h]
  \centering
  {\small{
  \begin{tabular}{@{}lc@{}}
    \toprule
    Methods & Accuracy \\
    \midrule
    Highway CNN \cite{highway} & 0.52 \\
    Tranfer learning VGG19 \cite{transfer}  & 0.57\\
    CNN-Transformer \cite{CNN_Trans} & 0.66\\
    \bottomrule
  \end{tabular}
  }}
  \caption{State-of-Art in offline learning.}
  \label{tab:SOA_batch}
\end{table}

\begin{table}[h]
  \centering
  {\small{
  \begin{tabular}{@{}lc@{}}
    \toprule
    Methods & Accuracy \\
    \midrule
    KD \cite{deepCL2022} & 0.74 \\
    Active Learning \cite{incactlearn2020}  & 0.85 \\
    Deep KD of basic features \cite{maiden2023complexfacialexpressionrecognition}  & 0.74\\
    \bottomrule
  \end{tabular}
  }}
  \caption{State-of-Art in continuous learning.}
  \label{tab:SOA_inc}
\end{table}

\subsection{Results of CIL with Neural Networks}
\label{sec:ExpeNN}

In this study, we used two different convolutional networks. The first one (CNN1) is handcrafted and shallow. The second one is MobileNet (MBN), a pre-trained deep network. We wanted to compare their behaviors in the context of incremental learning of complex emotions. We detail below the architectures of both networks.

CNN1: Conv-Conv-Pool-Conv-Conv-Pool-Conv-Pool-Dense128-Dense64-Softmax

MBN: Conv-DS1-DS2-DS3-DS4-DS5-Pool-Dense128-Dense64-Softmax

where DS$i$ stands for Depthwise Separable convolutions. We also used Batch Normalization and DropOut when needed.
Therefore, the total number of parameters is 166K (0.65MB) for CNN1 and 2,431K (9.3MB) for MBN.

For training, we made data augmentation and we used the Adam optimizer with an initial learning rate of $10^{-3}$, a categorical cross-entropy loss and early stopping on the cross-validation set. Each experiment was repeated ten times and results display the mean accuracy.

In the following, we compare, for both CNNs:
\begin{itemize}
    \item The T-SNE mapping of extracted features on 2D to see whether classes are easily separable or not.
    \item The performance of each expert in the task-aware scenario.
    \item The performance of the ensemble of experts in the task-agnostic scenario.
\end{itemize}

\begin{figure}[htbp]
    \centering
    \includegraphics[width=0.4\textwidth]{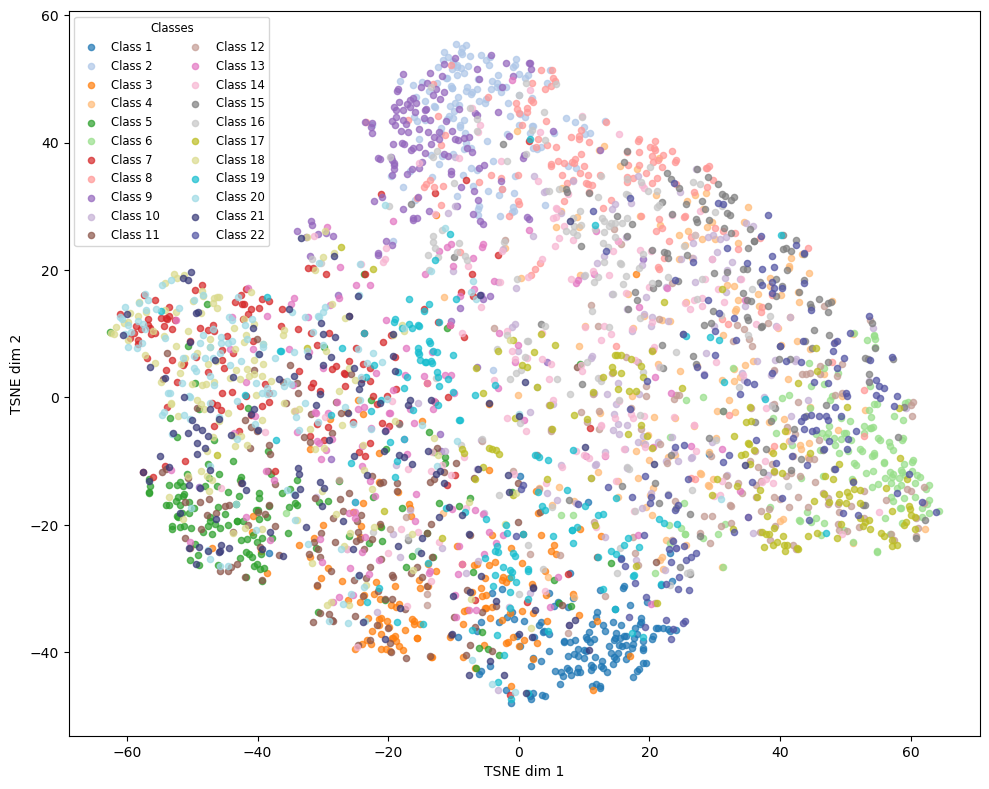}
    \caption{T-SNE projection with MBN as feature extractor.}
    \label{fig:MBN_TSNE}
\end{figure}

\medskip
\textbf{T-SNE mapping:}
Figure \ref{fig:MBN_TSNE} shows the T-SNE mapping of features when MBN is used as a features extractor. Since it is deeper than CNN1, it should extract better features. We can see that some classes  have a low within-class variance. Unfortunately, between-class variance is too low and most of the classes are overlapping.

\begin{figure}[htbp]
    \centering
    \includegraphics[width=0.4\textwidth]{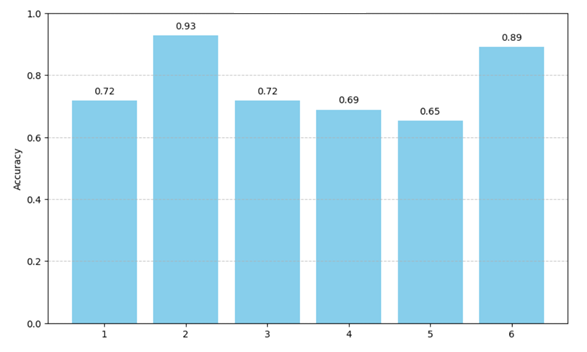}
    \caption{Task-aware performance using CNN1.}
    \label{fig:CNN1_task_aware}
\end{figure}

\begin{figure}[htbp]
    \centering
    \includegraphics[width=0.4\textwidth]{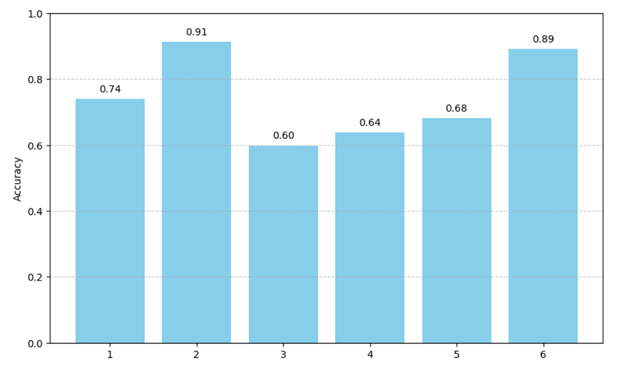}
    \caption{Task-aware performance using MBN.}
    \label{fig:MBN_task_aware}
\end{figure}

\medskip
\textbf{Task-aware scenario:}
Here, we report the performance of each expert on its own task. The classification of an input of task $t$ is done between the classes of this task only. We see in figures \ref{fig:CNN1_task_aware} and \ref{fig:MBN_task_aware} that task-aware accuracies vary between $0.6$ and $0.95$ for both networks. It seems that, whatever the complexity of the network, for this specific problem of facial emotion recognition,  performances are more or less the same. We can make the assumption that our convolutional networks have difficulties in finding optimal features for this problem. This poor performance is either due to a too simple architecture or a too small training set.

\begin{figure}[h]
    \centering
    \includegraphics[width=0.4\textwidth]{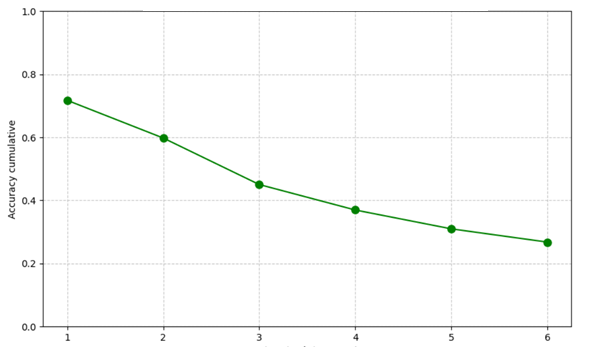}
    \caption{Incremental learning: Task-agnostic performance using CNN1.}
    \label{fig:CNN1_task_agnostic}
\end{figure}

\begin{figure}[h]
    \centering
    \includegraphics[width=0.4\textwidth]{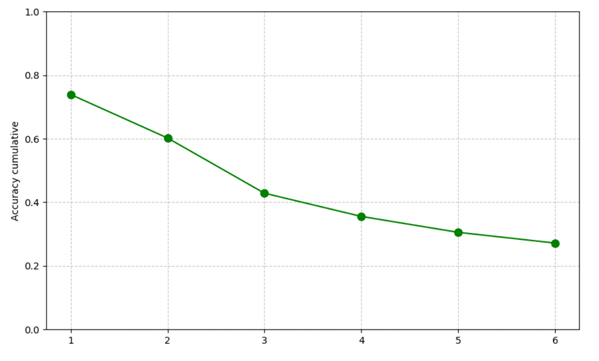}
    \caption{Incremental learning: Task-agnostic performance using MBN.}
    \label{fig:MBN_task_agnostic}
\end{figure}

\medskip
\textbf{Task-agnostic scenario:}
In this experiment, the task corresponding to the input is not known. Therefore, the classification must be made between all the classes learned. Figures \ref{fig:CNN1_task_agnostic} and \ref{fig:MBN_task_agnostic} show the cumulated task accuracy when $t$ task has been learned. We can see that both networks have the same behavior again. MBN is just slightly more accurate than CNN1 with an accuracy of 0.74 against 0.72 on task 1. But they perform equally with an accuracy of 0.27 on all tasks.

We make the assumption that accuracy decreases from task to task because task-dedicated networks learn features able to discriminate \textit{within} task classes. On the contrary, they do not learn features able to discriminate \textit{between} tasks. Since compound emotions are only slightly different from one to another, experts are not able to differentiate them. This may explain the difference observed between state-of-the-art results on CIFAR-100 detailed in \cite{rypeść2024} and those obtained in this work.

\subsection{Results of CIL with Action Units}

Now, we use directly AUs as feature vectors. Therefore, GMMs estimate the distribution in the AU space and compute a likelihood. The higher likelihood gives the predicted class.
We make the same experiments (mean accuracy on 10 runs) and visualization as in section \ref{sec:ExpeNN}.

\begin{figure}[htbp]
    \centering
    \includegraphics[width=0.4\textwidth]{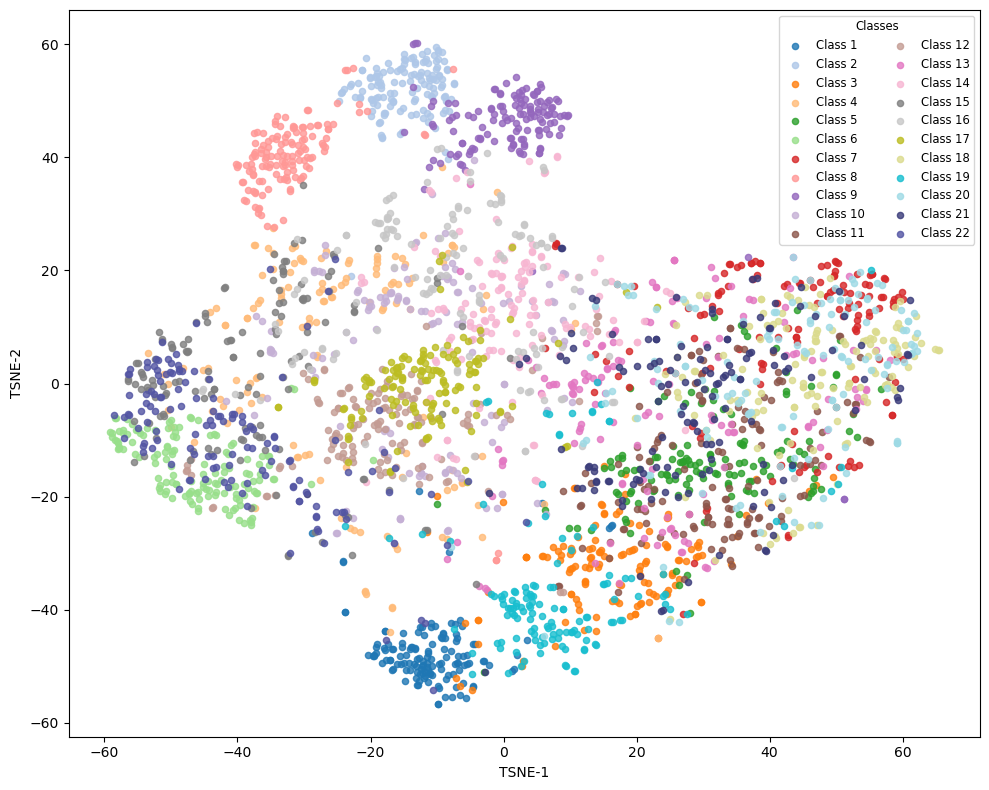}
    \caption{T-SNE projection with AUs.}
    \label{fig:AU_TSNE}
\end{figure}

\medskip
\textbf{T-SNE mapping:}
The T-SNE projection displayed on figure     \ref{fig:AU_TSNE} is better now. We clearly see more clusters corresponding to classes. But the overlap is still high.

\begin{figure}[h]
    \centering
    \includegraphics[width=0.4\textwidth]{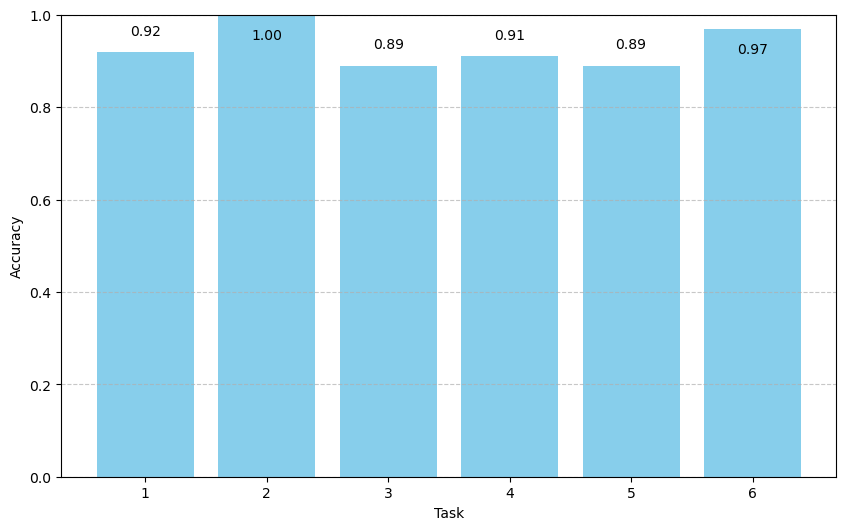}
    \caption{Task-aware performance using Action Units.}
    \label{fig:AU_task_aware}
\end{figure}

\medskip
\textbf{Task-aware scenario:}
We can see in figure \ref{fig:AU_task_aware} that task accuracies now vary between $0.88$ and $1.00$. This result is much more satisfactory, and mainly due to the use of Action Units as features since the rest of the ensemble has remained unchanged. These features are non-transient, so they do not vary when tasks change. Moreover, they provide rich semantic information about facial muscle activities that is quite sufficient to manage subtle differences between compound emotions.

\begin{figure}[h]
    \centering
    \includegraphics[width=0.4\textwidth]{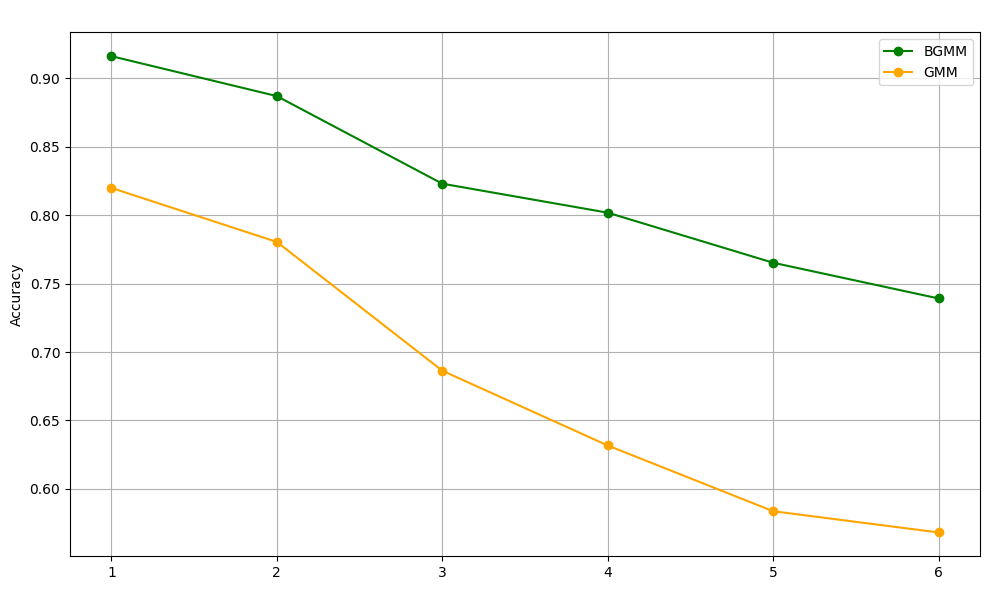}
    \caption{Task-agnostic performance using Action Units.}
    \label{fig:AU_task_agnostic}
\end{figure}

\begin{figure}[h]
    \centering
    \includegraphics[width=0.4\textwidth]{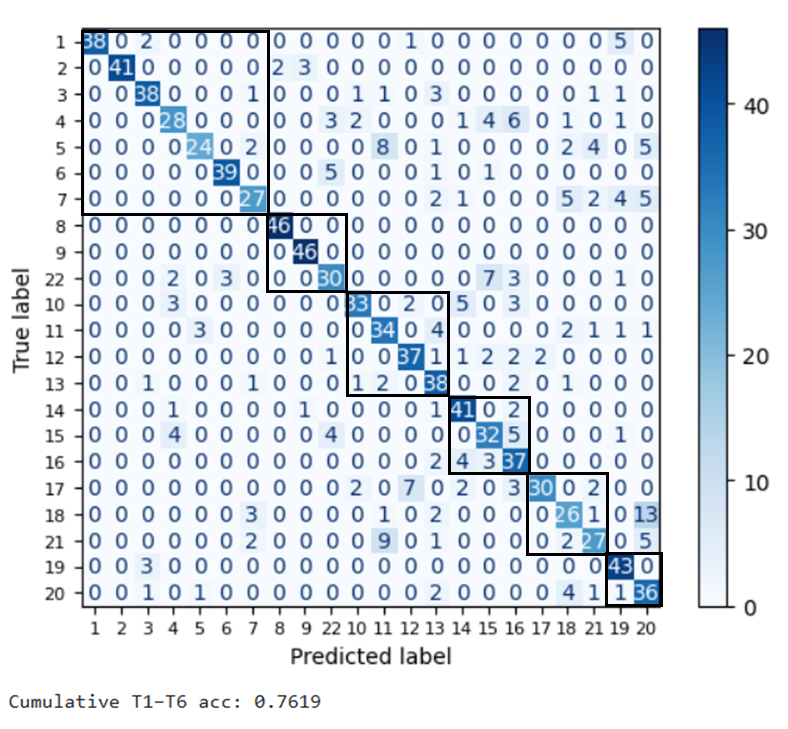}
    \caption{Task-agnostic confusion matrix.}
    \label{fig:conf_mat}
\end{figure}

\medskip
\textbf{Task-agnostic scenario:}
Results, presented in figure \ref{fig:AU_task_agnostic} are again much more convincing since accuracy evolves, from 0.92 (resp.0.82) for task 1 only to 0.75 (resp. 0.57) for all tasks learned incrementally with BGMMs (resp. GMMs). Notice that an accuracy of 0.75 is the state-of-the-art (see table   \ref{tab:SOA_batch}). 
Again, AU features clearly help to differentiate compound emotions.
The main drawback is that our classifier is purely generative since we only model underlying class distributions with class-conditional GMMs trained on the data of "their" classes. There is clearly a lack of a discriminative mechanism to effectively separate complex between-class boundaries. The confusion matrix, displayed in figure \ref{fig:conf_mat}, confirms that most confusions appear between classes of different tasks (bordered by black squares).

We can also compute the number of parameters of this model. Given $K$ the number of classes and $C$, the number of components $(\mu_c, \Sigma_c)$ per class, we have:

\begin{equation}
N_{param}= (S + \frac{S(S-1)}{2})*C*K    
\end{equation}

where $S$ is the dimension of the AU space.
Therefore, given $C=10$, for example, the number of parameters is about 0.03M to be added to the 6M parameters of OpenFace, which is quite low compared to vision transformers like ViT-B/16 which count 86M parameters. Thus, this sustainable solution, thanks to its low memory and carbon footprint, is very satisfactory.

\section{\uppercase{Conclusions}}
\label{sec:conclusion}

We have compared in this paper several feature extractors in order to learn incrementally complex, compound facial emotions.
We have shown experimentally that deep features, extracted by convolutional networks, were semantically too low to complete this task. We make the assumption that features extracted are not subtle enough to differentiate compound emotions.
On the contrary, using facial muscle activities leads to much better results, near state-of-the-art. This can be explained as these activities are different from one emotion to another and, above all, non-transient. In other words, they do not need to be continually learned.

In fact, these results are not that surprising. 
Psychological research shows it is easier  
for a human to recognize basic emotions or more complex
states like depression \cite{ekman2005} or pain \cite{pain2003} when AUs are primarily identified.
In a seminal work \cite{senechal_2014}, Senechal et al. demonstrate that recognizing emotions in the Action Unit space led to higher performances than recognizing emotions in a feature space extracted from images. This work is a new proof of this assumption.

We have also shown that complex Gaussian models provide an effective way to model class distributions, and that these distributions are sufficiently separated to allow successful emotion prediction, provided that the feature space is carefully chosen. This solution is not only efficient but also lightweight, with a minimal memory footprint, which makes it particularly suitable for embedded applications.

In the future, we will try to improve the system's accuracy by adding a second stage to the classifier, which aims at discriminating conflict between blended emotion classes. We will also apply this system in an unsupervised incremental way, to discover new emotion classes when they appear, annotate them using Active Learning or Vision Language Model and learn them.

\bibliographystyle{apalike}
{\small
\bibliography{egbib}
}

\end{document}